\documentclass{article}

\usepackage[final]{cpal_2024}

\usepackage{url}
\usepackage{pifont}
\usepackage{booktabs}
\usepackage{subfigure}
\usepackage{algorithm}
\usepackage{algorithmic}
\usepackage{balance}
\usepackage{amsmath}
\usepackage{wrapfig}
\usepackage{graphicx}

\newcommand{\mname}{\texttt{M2Cache}}

\title{Harnessing Your DRAM and SSD for Sustainable and Accessible LLM Inference with Mixed-Precision and Multi-level Caching}

\author{%
  Jie Peng\textsuperscript{1}\thanks{co-author}, ~Zhang Cao\textsuperscript{2}\footnotemark[1],~Huaizhi Qu\textsuperscript{3},~Zhenyu Zhang\textsuperscript{2},~Chang Guo\textsuperscript{2},\\~Yanyong Zhang\textsuperscript{1},~Zhichao Cao\textsuperscript{2},~Tianlong Chen\textsuperscript{3} \\
  \textsuperscript{1}University of Science and Technology of China, \textsuperscript{2}Arizona State University,\textsuperscript{3}University of North Carolina at Chapel Hill\\
  \texttt{pengjieb@mail.ustc.edu.cn, yanyongz@ustc.edu.cn\\ \{zcao59, zzhan641, cguo51, zhichao.cao\}@asu.edu.cn,\\ \{qhz991029,tianlong\}@cs.unc.edu,}
}

\begin{document}

\maketitle

\begin{abstract}
Although Large Language Models (LLMs) have demonstrated remarkable capabilities, their massive parameter counts and associated extensive computing make LLMs' deployment the main part of carbon emission from nowadays AI applications. Compared to modern GPUs like H$100$, it would be significantly carbon-sustainable if we could leverage old-fashioned GPUs such as M$40$ (as shown in Figure~\ref{fig:tisser}, M$40$ only has one third carbon emission of H$100$'s) for LLM servings. 
However, the limited High Bandwidth Memory (HBM) available on such GPU often cannot support the loading of LLMs due to the gigantic model size and intermediate activation data, making their serving challenging. For instance, a LLaMA2 model with $70$B parameters typically requires $128$GB for inference, which substantially surpasses $24$GB HBM in a $3090$ GPU and remains infeasible even considering the additional $64$GB DRAM. To address this challenge, this paper proposes a mixed-precision\footnote{The precision denotes the numerical precision like FP16, INT8, INT4.} and multi-level caching (\mname) with a model modularization algorithm to enable LLM inference on outdated hardware with resource constraints.

Specifically, our \mname~first modulizes neurons in LLM and creates their importance ranking. Then, it adopts a dynamic sparse mixed-precision quantization mechanism in weight space to reduce computational demands and communication overhead at each decoding step. It collectively lowers the operational carbon emissions associated with LLM inference. Moreover,~\mname~introduces a three-level cache management system with HBM, DRAM and SSDs that complements the dynamic sparse mixed-precision inference. To enhance communication efficiency, \mname~maintains a neuron-level mixed-precision LRU cache in HBM, a larger layer-aware cache in DRAM, and a full model in SSD. The cache management uses layer-wise pre-loading from SSD to DRAM, and asynchronous loading from DRAM to HBM to overlap the HBM cache miss with the GPU computation for performance. With the decreased communication overhead and the usage of SSDs, the operational carbon footprint of LLMs inference is further reduced. Extensive experimental results demonstrate the effectiveness of our \mname. Compared with DeepSpeed Zero-Infinity, \mname~significantly improves the token generation speed by up to $\times 10.51$. carbon emission reduction up to $\times 7.67$. 
\end{abstract}

\section{Introduction}

Global warming and climate change increasingly threaten our economy and society~\cite{abbass2022review, ma2023crisis},
this urgency has spurred a shift towards sustainable investing, particularly in green computing, with a focus on minimizing carbon emissions~\cite{2024preventing, vartziotis2024carbon}.
Concurrently, the rapid advancement and deployment of Large Language Models (LLMs)~\cite{vicuna2023, driess2023palme, chung2022scaling, zhang2022opt, chowdhery2022palm, sanh2022multitask, wei2021finetuned, gpt4, dubey2024llama3, anil2023gemini, reid2024gemini, team2024gemma, dai2024deepseekmoe, jiang2023mistral, bai2023qwen, yang2024qwen2} have demonstrated remarkable capabilities across a wide range of tasks, including natural language processing~\cite{wei2021finetuned, workshop2023bloom, zeng2023glm130b, rafailov2024direct}, question answering~\cite{nakano2021webgpt, siriwardhana2023improving}, and code generation~\cite{chen2021evaluating, ni2023lever, hong2023metagpt, muennighoff2024octopack}. 
Therefore, the LLMs deployment is becoming a growing factor for AI’s
overall environmental impact.
This issue poses a substantial challenge to the goal of advancing sustainable AI development, emphasizing the need for solutions that address the environmental impact of LLM deployment~\cite{wu2022sustainablea}. For example, ChatGPT's API call emitted approximately $24.24$ $ktCO2e$ in January 2023~\cite{daramon2024assessing}.
This is equivalent to the amount of carbon dioxide absorbed by $12.12$ million trees in one year.

\begin{wrapfigure}{r}{0.45\textwidth}
  \centering 
  \vspace{-4mm}
  \includegraphics[width=\linewidth]{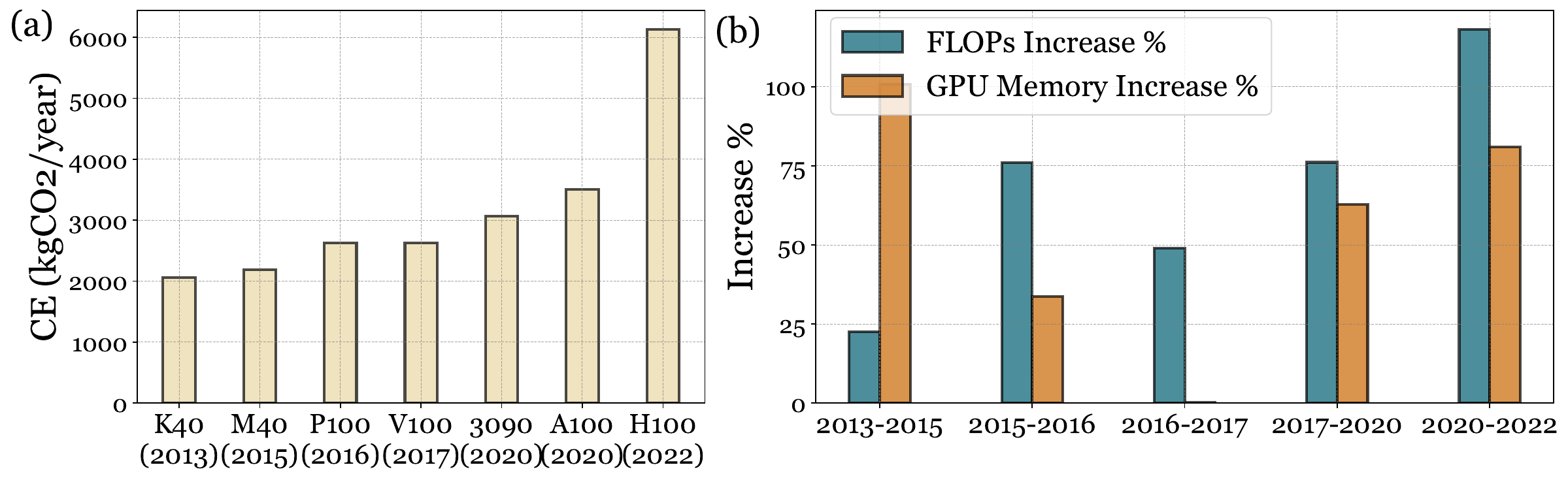}
    \vspace{-4mm}
  \caption{\textit{The operational carbon emission (CE)~\cite{2024review}, Flops, and GPU memory change over the years.} Over the past decade, operational carbon emissions, FLOPs, and memory of GPUs have consistently increased. However, the growth rate in FLOPs has outpaced that of GPU memory in almost every product iteration.}
  \label{fig:tisser} 
\end{wrapfigure}

The deployment of LLMs reaches a crucial point during the inference phase, where AI functionalities become accessible to users. 
This phase is particularly demanding in terms of resources, a consequence of both the substantial size of these models~\cite{wu2022sustainablea} and the need to meet high standards for service quality and responsiveness~\cite{wilkins2024hybrid}. 
Thus, making LLM inferences with better tradeoffs between sustainability and service quality (accuracy and efficiency) is a key step towards aligning AI development with environmental goals~\cite{nocode.aifuture}.

To reduce the carbon emissions for LLM services, leveraging old-fashioned hardware for LLM inference emerges as a promising strategy. 
For instance, figure~\ref{fig:tisser} shows that old-fashioned GPUs have much smaller carbon emissions compared with cutting-edge devices like A100 and H100.
In addition, these devices reduce the embodied carbon and the need for new hardware production and consequently, the generation of electronic waste~\cite{2023turning, switzer2023junkyard}. 
Finally, it does not significantly impair the balance between LLM accuracy and efficiency.
Current hardware, already in circulation, has been shown to possess adequate computing capabilities for conducting effective LLM inference without compromising on accuracy or efficiency~\cite{wilkins2024offlineb, samsi2023words}. 
For example, the K40 has 5.04 TFLOPs computation ability and LLaMA-7B requires around 19.61 GFLOPs to generate one token~\cite{lin2024slim}.
As we showed in Figure~\ref{fig:tisser}, the computational power of the device is sufficient for LLMs inference, even if it was released 10 years ago.

However, LLM inference requires large High-Bandwidth Memory (HBM) capacities to load LLM parameters and cache intermediate data (e.g., key-value cache).
Therefore, the limited HBMs of old-fashioned hardware will become the bottleneck for LLMs deployment~\cite{alizadeh2024llm, song2023powerinfer}.
For instance, the older device Nvidia V100 GPU, includes 32GB of HBM, which is insufficient for loading LLMs with more than 13 billion float-point 16 parameters.
In contrast, the newer and top-tier ones like H100 GPU feature 80 GB HBM, which can handle LLMs with larger than 33 billion float-point 16 parameters.

There are existing works that can partially alleviate the limitations of HBM in old-fashioned GPUs, such as pruning~\cite{sun2023simple, zhang2024h2o, xiaShearedLLaMAAccelerating2024}, quantization~\cite{nagel2021white, gholami2021surveyquantizationmethodsefficient, frantar2022gptq, lin2024awq, dettmers2022llmint88bitmatrixmultiplication}, and optimizing Key-Value (KV) Cache usage~~\cite{zhang2023_2, liu2024intactkv}. 
However, KV cache optimization can not address the situation in which HBM is not even enough to load model parameters.
Pruning and quantization can reduce the HBM demand for loading LLM model parameters.
Nonetheless, these methods introduce the challenge we defined as \textbf{Parameter Over-correction}: To accommodate larger models within constrained HBM capacities, parameters must be compressed to a very low bit or significantly pruned.
This issue becomes worse as model sizes increase without a corresponding expansion in HBM capacity.

Other alternative solutions involve offloading LLM parameters to either CPU DRAM or high-performance SSDs, as explored in recent research~\cite{sheng2023flexgen, alizadeh2024llm, yuan2024efficient,lee2024infinigen,pan2024instinfer}.
This method leverages the observation that not all neurons\footnote{The neuron is defined as a specific row/column in a weight matrix} of the LLM are active at once during inference. Consequently, it temporarily moves certain components to slower, but higher-capacity memory/storage options like DRAM and high-performance SSDs.
However, this strategy encounters a significant challenge known as \textbf{Bandwidth Overwhelming}: The communication between HBM and other memory/storage is overwhelmed by the frequent module loading to HBM, becoming a bottleneck for LLM inference efficiency.
For example, prevailing HBM hardware uses PCIe interfaces with bandwidths below $64$ GB/s. 
The maximum inference speed under this bandwidth of LLaMA-13B is $4$ token/s when feed-forward network (FFN) parameters are offloaded to DRAM.
The root cause of this problem lies in the substantial data volume of the modules being offloaded~\cite{sheng2023flexgen} and the frequent need to reload these modules back to the GPU~\cite{alizadeh2024llm}. 
Such behaviors saturate the communication bandwidth between HBM and DRAM, leading to inferencing speed bottlenecks. Also, the DRAM space may not be enough to cache the whole model, and a portion of the space is used/reserved for other applications (e.g., databases). The available DRAM space can be limited to old-fashioned servers. Moreover, the frequent data movement, DRAM access, and large volume of DRAM requirements also cause extra carbon emissions.

To address these fundamental challenges, we propose a sustainable and cost-effective LLM inference architecture with \textbf{M}ix-precision and cost-effective \textbf{M}ulti-level \textbf{cache} (called \textbf{~\mname}) over old-fashioned computing servers with HBM-limited GPUs.
\mname~consists of two fundamental novel designs:~\textit{dynamic sparse mixed-precision inference}, and~\textit{predicting-driven multi-level cache as HBM extension}.
It harmonizes quantization and offloading methods in LLMs and combines them with a customized multi-level cache to enhance memory efficiency and carbon emissions.

The \textbf{dynamic sparse mixed-precision inference} operates within the FFNs of LLMs, treating each row in the first layer of an FFN and the corresponding column in subsequent layers as a neuron.
Active neurons, identified by predictors specific to each layer based on its input, are selectively transferred from DRAM to GPU memory, a process termed ``dynamic sparsity''.
Then it quantizes the less important part in active neurons to lower bits and uses these mixed-precision active neurons to do the computing in GPU.
The inactive neurons in GPU memory will be offloaded to DRAM to save GPU memory and reduce embodied carbons from HBM.
The process involves three steps: 1) \textit{Active Neuron Identification}: A low-rank predictor locates the necessary neurons for specific text generation tasks. 2) \textit{Selective Loading into GPU}: Only the identified active neurons are loaded, optimizing memory use, and 3) \textit{Active Score-based Quantization}: Neurons with lower active scores are quantized to a smaller number of bits, conserving HBM.
Loading these quantized neurons saves the communication bandwidth, due to the data volume being smaller compared to high-precision neurons, thereby lessening the bandwidth overwhelming challenge,
By choosing the right ratio of neurons of different precisions, we can have more neurons while maintaining the precision of critical neurons, thus alleviating the \textbf{Parameter Over-correction} challenge.
Our method saves computation by utilizing a subset of neurons during inference.
These reductions lower carbon emissions during LLM inference, thus improving the sustainability of LLM deployment.

The \textbf{predicting-driven multi-level cache} is designed to efficiently manage neuron data across three types of memory/storage: GPU HBM, DRAM, and SSD, which utilizes DRAM and SSD as the extension for GPU HBM.
The power consumption, carbon emissions, price, and speed of these storage media are from high to low.
Specifically, the SSDs are used to cache all FFN parameters. This allows for a large capacity of data to be held at the lowest cost and carbon emission.
The DRAM maintains a layer-aware FIFO queue, it loads multiple to-be-used FFNs from SSDs and helps in managing the slower access speed of SSDs.
In the GPU memory, we design the neuron-level management for each LLM layer, which retains the most frequently accessed active neurons to ensure quick retrieval.
More importantly, this multi-level cache complements the dynamic sparse mixed-precision inference with a two-level caching strategy: 1) \textit{GPU-DRAM Cache}: Utilizing an LRU cache mechanism, this level stores frequently accessed active neurons directly in the GPU cache.
The active neurons are identified by the predictor in the dynamic sparse mixed-precision inference.
2) \textit{DRAM-SSD Cache}: Given the comparatively slower SSD bandwidth, a proactive pre-loading policy is employed. This anticipates and pre-loads neurons likely to be needed soon from SSDs to DRAM, further alleviating DRAM and GPU memory constraints.
The GPU-DRAM Cache significantly lowers the demand for loading neurons into GPU memory, effectively reducing GPU-CPU bandwidth usage and addressing the overwhelming bandwidth challenge. 
The reduced GPU-CPU bandwidth and the utilization of SSDs for all parameters loading decrease the power consumption of LLM inference.

We evaluate~\mname~across different models and previous generations of hardware: LLaMA-7B, LLaMA-13B, LLaMA-70B, and Falcon-40B on a single GeForce RTX 3090. 
Our machine is equipped with 64 GB of DRAM and a 1 TB SSD that uses the PCIe 3.0x4 interface. It runs on Ubuntu 22.04 and features an AMD 6950x CPU. To simulate a scenario with limited CPU resources, we utilize only one core of the CPU for mix-precision and cache management.
Compared with the state-of-the-art offloading framework DeepSpeed Zero-Infinity \cite{2022zeroinference}, for LLaMA-7B, ~\mname~ reduces inference latency by up to $\times 7$. Similarly for LLaMA-13B, we achieve up to $\times 14$ inference speed up. ~\mname~ also enables LLaMA-70B and Falcon-40B, resulting in up to 0.3835 and 0.312 tokens/s speed up on a single GeForce RTX 3090. For carbon emissions, ~\mname~ achieves up to $\times 7$ reduction compared with Zero-Infinity. The dynamic sparse mixed-precision inference achieves about $\times 1.47$ inference latency reduction and $\times 1.45$ carbon emission reduction. The prediction-driven multi-level cache achieves about $\times 2.15$ inference latency reduction and $\times 2.17$ carbon emission reduction.
The main contribution of our paper includes:
\begin{itemize}
    \item \textbf{In-depth Analysis of Inference Overhead.} We investigate the inference overhead of LLMs in memory-limited devices and identify the challenges associated with deploying LLM by quantization and offloading.
    \item \textbf{Dynamic Sparse Mixed-Precision Quantization.} We leverage the neuron scores in dynamic sparse inference to introduce mixed-precision quantization in LLMs to improve the performance of LLMs in limited memory resources scenarios.
    \item \textbf{Innovative Multi-Level Cache System.} We introduce multi-level cache, including GPU-DRAM and DRAM-SSD cache, to utilize the design space opened by the dynamic sparse mixed-precision quantization.
    Our multi-level cache further adequately enhances the inference performance.
    \item \textbf{Better Sustainability.} To our best knowledge, ~\mname~is the first research that addresses LLM inferencing sustainability with old-fashioned hardware. Moreover, \mname~improve the sustainability of LLMs without accuracy sacrifice.
\end{itemize}

\section{Background}
\subsection{LLM Inference}

\smallskip\noindent\textbf{Prefiling and Decoding.}
By default, the parameter of LLM and the medium result are stored in the GPU memory during inference.
The process begins with the prefill phase, where the system processes the user's input prompt to generate the initial output token~\cite{gpt1, gpt2, gpt3, gpt4, touvron2023llama2}. 
Following this, the decode phase takes over, producing subsequent output tokens sequentially. 
In this phase, each newly generated token is used by the model in GPU to create the next one. 
This process continues until the model produces a special token indicating the end of the sequence. 

Notably, for each request, the LLM does one full forward pass through the LLM for prefiling and multiple full forward passes for decoding.
Each full forward pass of decoding only processes the new token generated in the previous step.
This phase results in the underutilization of computation resources, as the process becomes primarily constrained by memory~\cite{liu2023deja, sheng2023flexgen, li2024llm, zhang2024h2o}.
Memory consumption during this process is notably affected by two main factors: the Key-Value (KV) Cache~\cite{sheng2023flexgen, zhang2024h2o, kwon2023efficient, kwon2023efficient} and the parameters from FeedForward Networks (FFNs)~\cite{liu2023deja, song2023powerinfer}.
The KV cache optimization has many mature solutions, like storing it in the CPU~\cite{sheng2023flexgen} or pruning redundant cache blocks~\cite{zhang2024h2o}.
The FFNs is another major reason why LLMs take large GPU memory.
It accounts for a significant portion of LLM parameters, consuming over half of the GPU memory space. For instance, FFNs consume from $63.99\%$ of parameters in the LLaMA-7B model to $72.41\%$ in the LLaMA-70B model.
Investigating methods to streamline FFN parameters is one of the pathways to more efficient and sustainable LLM deployments.

\smallskip\noindent\textbf{Dynamic Sparse Inference.}
Sparsity is a natural approach to reducing computation and memory costs during neural network inference by discarding less important weights\cite{molchanov2016pruning, liu2018rethinking, hoefler2021sparsity}. 
In the context of LLMs, many studies focus on leveraging sparsity to achieve memory savings and lower computation\cite{frantar2023sparsegpt, bansal2022rethinking, liu2023deja}. 
One important type of sparsity is the dynamic contextual sparsity introduced in Deja Vu~\cite{liu2023deja}.
The dynamic contextual sparsity in FFNs activates a subset of parameters in FFNs for each new token during the decoding phase without hurting model accuracy.
It opens a new pathway for memory optimization: by predicting which subsets of the model will be inactive for a specific input, these segments can be offloaded from GPU memory to more abundant storage solutions like DRAM or SSDs. 
This process, informed by the Deja Vu predictor, allows for significant GPU memory savings while maintaining the model's output quality~\cite{liu2023deja}.

\smallskip\noindent\textbf{LLM Quantization.}
Weight quantization is a widely adopted method for model compression to reduce the storage costs of LLMs~\cite{nagel2021white, gholami2021surveyquantizationmethodsefficient, frantar2022gptq, lin2024awq, dettmers2022llmint88bitmatrixmultiplication}. 
In LLM quantization, weights and activations are stored with fewer bits at a lower numerical precision.
These methods decrease the memory required for each parameter.
The major targets to be quantized are the weights of linear layers ($i.e.$, matrix multiplication), which account for more than $99\%$ of the overall LLM weights~\cite{frantar2022gptq, lin2024awq, xiao2023smoothquanta, dettmers2023case}.
Besides, the process of ``dequantization'' refers to transforming the quantized weights back to FP16.

\subsection{GPU servers for LLM inference}\label{sec: gpu_server}

\smallskip\noindent\textbf{Old-fashioned GPUs.} Old-fashioned GPUs are typically designed or modified for specialized tasks, such as gaming, content creation, or specific AI workloads. These GPUs may have slight modifications based on user requirements or optimizations for particular applications. Some examples include gaming GPUs repurposed for AI inference. Custom GPUs usually have lower HBM capacities, around 4-24 GB. For example, NVIDIA GeForce RTX 3090 have 24GB HBM, NVIDIA GeForce RTX 3060 have 12GB HBM. They are typically more universal within consumer markets and are more commonly used than top-tier GPUs \cite{song2023powerinfer}.

\smallskip\noindent\textbf{Top-tier GPUs} Top-tier GPUs are built for high-performance computing (HPC), large-scale AI training, inference, data centers, and scientific research. These GPUs focus on reliability, performance, and scalability, often featuring much larger memory capacities. Its HBM sizes range from 40 GB to 80 GB and beyond. For example, NVIDIA’s A100 has 40GB or 80GB HBM size. Enterprise GPUs are optimized for specific professional and research tasks, making them less universal. These GPUs are designed for server environments, often with support for multi-GPU configurations, data center integration, and industry-specific software stacks.

\smallskip\noindent\textbf{Carbon footprint of servers.} The carbon footprint is a metric for quantifying the amount of greenhouse gas emissions ($gCO_2$) generated. When executing LLMs inferencing on a server, its carbon footprint comprises the \textit{embodied carbon emissions} (denoted as \textbf{\textit{ECE}}) and \textit{operational carbon emissions} (denoted as \textbf{\textit{OCE}}). As shown in Formula \ref{formula: carbon_emission}, the carbon footprint is the sum of embodied and operational carbon emissions. Embodied carbon emissions represent the carbon emissions associated with the manufacturing and packaging of computer components, effectively “embodied” within the device itself. For an inference request processed by a server, its share of embodied carbon is proportional to the execution time relative to the device’s overall operational lifespan. Using old-fashioned hardware (e.g., most of the GPU servers may already configured with GeForce RTX 3090) instead of investing in the latest hardware (e.g., H100) for LLM inferencing can explicitly reduce the embodied carbon.

The operational carbon emissions come from the electricity grid that powers the server, which powers the hardware (e.g., GPUs, CPUs, DRAM, and SSDs) that serves the inference. The carbon intensity (denoted as $CO^{Intensity}_2$) of the grid, representing the amount of CO2 emission per energy usage (gCO2/kWh), reflects the “greenness” of the energy source. For example, wind turbines have lower carbon intensity than coal power plants. Due to the difference in availability and stability of renewable energy, carbon intensity varies significantly over time and across geographical regions. The carbon intensity is the multiplier to the request’s energy consumption when quantifying its operational carbon footprint. Usually, the general per-bit carbon emission (including embodied and operational) of HBM, DRAM, and NAND-flash-based SSDs are from high to low \cite{wang2024designing, gupta2022act}. As shown in Formula \ref{formula: carbon_emission}, OCE is related to the consumed energy, and energy is related to the occupied memory and runtime. The total carbon footprint (donates as \textbf{CF}) is the sum of OCE and ECE. 

\begin{equation} \label{formula: carbon_emission}
\left\{
\begin{array}{l}
\text{CF} = \text{OCE} + \text{ECE} \\
\text{OCE} = \text{Energy} \times \text{Energy\_to\_Carbon} \\
\text{Energy} = \text{Memory\_Size} \times \text{Power\_Unit} \times \text{Runtime} \\
\end{array}
\right.
\end{equation}

\section{Motivation}

The increasing deployment of LLMs has significantly raised carbon emissions during inference, driven by the growing model sizes. As a result, improving the efficiency and sustainability of LLM inference is becoming increasingly crucial~\cite{298775}. This section provides an in-depth analysis of the factors contributing to the rising carbon footprint of LLM inference and investigates the potential of leveraging old-fashioned GPU servers to reduce these emissions.

\subsection{The dilemma between LLM inference demands and more carbon emissions}\label{sec_motivation_1}

As LLMs continue to evolve, their parameter counts have grown substantially, now ranging from several billion to hundreds of billions. Larger models are generally preferred for inference due to their improved performance, particularly in terms of accuracy. For instance, a 70B general-purpose model demonstrates significantly higher accuracy compared to a 7B model without any specific fine-tuning \cite{touvron2023llama, touvron2023llama2, dubey2024llama3, zhang2022opt}. 
However, inferencing larger models requires advanced GPU servers, primarily due to the increased High HBM requirements. Among the various GPU specifications, HBM size is the most critical factor for LLM inference ~\cite{alizadeh2024llm, song2023powerinfer}, driving the need for advanced GPUs with larger memory capacities. As the size of the models increases, the corresponding HBM requirements also rise, necessitating the use of more capable GPUs. For example, old-fashioned GPUs like the RTX 3090 and RTX 4090 are only sufficient for 7B models before exhausting their memory, whereas more advanced GPUs such as the A100 and H100 can support models with up to 13B and 33B parameters, respectively, offering substantially enhanced performance. This growing demand for high-performance GPUs underscores the critical need for more advanced GPUs to support the evolving landscape of LLM inference.

The scarcity of advanced GPUs, coupled with surging demand, has accelerated the production of newer, more sophisticated GPU models. However, the manufacturing of these cutting-edge GPUs incurs substantial embodied carbon emissions. For example, the embodied carbon footprint of an A100 GPU is approximately 150 kg CO2, equivalent to driving a conventional fuel-powered vehicle for 600–800 kilometers \cite{luccioni2023estimating}. As GPU production scales to meet the increasing computational demands of large language models (LLMs), the associated carbon footprint also escalates, exacerbating environmental impacts and raising sustainability concerns. The persistent cycle of developing larger models, requiring more advanced hardware, and generating higher carbon emissions underscores the critical need for sustainable approaches to LLM inference and deployment.

\subsection{Opportunities in existing old-fashioned servers}

As discussed in Section~\ref{sec: gpu_server}, old-fashioned GPU servers, such as those equipped with RTX 3090 GPUs, typically have limited High Bandwidth Memory (HBM) sizes along with a certain amount of DRAM and SSD storage. This section examines the potential of these old-fashioned servers to lower the financial and environmental costs of large language model (LLM) inference, thereby making LLM deployment more feasible and broadly accessible.

\noindent \textbf{Opportunity 1: } \textit{The carbon emissions of inferencing on these existing old-fashioned GPUs are much smaller than the top-tier GPUs.}

As described in Section \ref{sec: gpu_server}, the carbon footprint of GPUs includes both embodied and operational emissions. Embodied emissions arise during the manufacturing process of the hardware. In this study, we leverage existing old-fashioned GPUs that are already deployed and in use, thereby incurring no additional embodied emissions. Conversely, top-tier GPUs are often newly manufactured and less readily available, contributing to significant embodied carbon costs. Thus, performing inference on existing old-fashioned GPUs avoids additional embodied carbon emissions.

Figure \ref{fig:tisser} shows the operational carbon emissions of different types of GPUs. As shown in Figure \ref{fig:tisser}, we observe that for top-tier GPUs like the A100 and V100, their operational carbon emissions are very close to those of old-fashioned GPUs like the RTX 3090. This indicates that there is no significant improvement in carbon emissions when enterprise GPUs are involved. Thus, performing inference on existing old-fashioned GPUs doesn't result in much additional operational carbon emissions, and in some cases, it can even reduce carbon emissions.

\noindent \textbf{Opportunity 2: } \textit{Inference on existing old-fashioned servers is cheaper and remove the inference barrier .}

Top-tier GPUs exhibit significantly higher price points compared to old-fashioned counterparts. Notably, the NVIDIA A100 GPU is priced approximately ten times higher than the Geforce RTX 3090 and five times higher than the Geforce RTX 4090, illustrating a substantial cost disparity \cite{super}. The H100, a more advanced modern GPU, commands an even higher premium. This pronounced disparity highlights the significant cost impact of deploying top-tier hardware in LLM inference scenarios.

As discussed in Section \ref{sec: gpu_server}, old-fashioned GPUs are more widely used than top-tier GPUs and they are deployed in most data centers, institutions, organizations, and even personal workstations. Therefore, if we can perform large model inference on these old-fashioned GPUs, it could remove the LLM inference barrier for most GPU servers and make LLM inference more universally accessible.

\section{Challenges}\label{sec_challenge}

Although utilizing existing old-fashioned servers for LLM inference is more sustainable the limited HBM of these servers poses several challenges to running large LLM inference (e.g., LLaMA-2 13B and even 70B). Due to the limited HBM, the model size can not fit into GPU. 

\begin{figure*}[t]
\centering
\includegraphics[width=\linewidth]{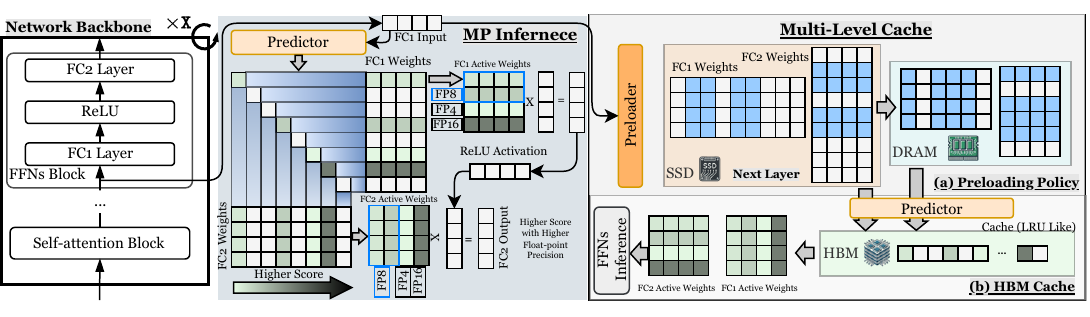}
\vspace{-6mm}
\caption{Overall Architecture of~\mname.}
\label{fig: overall_arch}
\end{figure*}

First, the CPU computation power could increase the inference latency if we rely on the CPU computation power.
CPUs, while versatile, may not possess sufficient computational speed for efficient LLM inference, resulting in slower response times. 
Then the challenge moves to the deployment of computation tasks on high-bandwidth memory (HBM) devices, such as GPUs.
Although GPUs in existing devices are capable of handling complex computations efficiently, their relatively limited HBM capacity becomes a critical bottleneck~\cite{alizadeh2024llm, song2023powerinfer}.
For instance, while older devices like the V100 GPU offer 32 GB of HBM, newer and top-tier ones like the A100 GPU provide 80 GB. 

There are existing works that can partially alleviate the limitations of HBM, such as quantizing MLP (Multi-Layer Perceptron) weights and optimizing Key-Value (KV) Cache usage~~\cite{zhang2023_2, liu2024intactkv}. 
These methods aim to decrease the memory required for each parameter, effectively saving HBM space. 
However, these approaches introduce a new challenge known as \textbf{Parameter Over-correction}. 
To accommodate larger models within constrained HBM capacities, parameters are often compressed to very low bit sizes.
Although this saves space, it can also lead to a significant reduction in the accuracy of the models~\cite{yao2023zeroquantv2}. 
This issue becomes worse as model sizes increase without a corresponding expansion in HBM capacity.

Another alternative strategy for managing LLM inference involves offloading LLM parameters to either CPU DRAM or high-performance SSDs, as explored in recent research~\cite{sheng2023flexgen, alizadeh2024llm}. 
This method leverages the observation that not all parts of the LLM are active at once during inference. Consequently, it temporarily moves certain components to slower, but higher-capacity storage options like DRAM and high-performance SSDs.
However, this strategy encounters a significant challenge known as \textbf{Bandwidth Overwhelming}. The root of this problem lies in the substantial data volume of the modules being offloaded~\cite{sheng2023flexgen}, or the frequent need to reload these modules back to the GPU~\cite{alizadeh2024llm}. 
Such behaviors saturate the communication bandwidth between High Bandwidth Memory (HBM) and DRAM, leading to bottlenecks. 
As a result, the efficiency of LLM inference is compromised, as the system struggles to load the necessary LLM modules from DRAM to HBM in a timely manner.

\section{\mname~Designs}

Currently, some studies focus on KV cache offloading. For instance, AttentionScore~\cite{gao2024costefficient} leverages DRAM and SSD storage to store the generated KV cache for reuse in subsequent conversations. However, old-fashioned GPUs often lack sufficient HBM to store even the model weights, let alone the KV cache. Therefore, this paper focuses on more aggressive model weight offloading, aiming to enable inference of larger models on old-fashioned servers. Additionally, these servers typically have low-performance CPUs, and their computational resources are frequently occupied by other programs~\cite{alsubaihi2017runtime}. Consequently, this study concentrates on GPU-centric offloading~\cite{yamato2023proposal}, relying exclusively on the GPU for computation during LLM inference.

Based on the above analysis, we are aiming to make it possible for customer-level old-fashioned servers to infer larger models (e.g., LLaMA 2 70B model) to meet performance requirements by offloading model weights to the host side so that LLM inference can be more sustainable, cheaper, and universally accessible.

\subsection{\mname~Design Overview}

We first present the design overview of~\mname, a model and cache co-design that provides sustainable and accessible LLM inference on old-fashioned servers equipped with GPU, CPU, limited DRAM, and SSDs. as shown in Figure~\ref{fig: overall_arch}.
It consists of two key techniques: \textit{dynamic sparse mixed-precision inference} (\texttt{MP Inference}) and \textit{multi-level cache}.
The \textit{dynamic sparse mixed-precision inference} uses a predictor to estimate active neurons for the current input inference.
According to the estimated score, active neurons can be split into different precisions.
The higher the score, the higher the precision of the neuron.
We will offload inactive neurons into DRAM to save HBM space and open the design space for advanced multi-level cache management.
The \textit{multi-level cache} complements \texttt{MP Inference}. It first utilizes a preloader and a two-level DRAM cache-based tiered-cache to hide SSD as the bottom tier, creating the effect of "infinite" DRAM. Second, it employs a high-performance model layer-based HBM cache to reduce the time spent loading neurons from DRAM.

To address the over quantization challenge, the \texttt{MP Inference} employs mixed precision neurons which allows us to keep the critical neurons with higher precision alongside these low-precision neurons.
These high-precision critical neurons underpin the LLM accuracy. (Section~\ref{mp_inference}.) To resolve the bandwidth overwhelming issues, the \textit{multi-level cache} includes a GPU-managed LRU cache that can reduce the number of neurons required to load from DRAM to GPU memory. (Section~\ref{hbm_cache} and Section~\ref{preload_policy}.) Moreover, to address the space limited of HBM ($e.g.$, the GPU memory), \textit{multi-level cache} involves the SSDs that can enable LLM inference when HBM and DRAM together are not enough to load the whole model parameters.
The corresponding synchronous layer-wise DRAM preloading policy diminishes the effect of low bandwidth of SSDs during inference. (Section~\ref{preload_policy}.)

Carbon emission reductions are achieved through both \texttt{MP Inference} and \textit{multi-level cache} strategies. \texttt{MP Inference} decreases computational carbon by using only a subset of neurons during inference. The \textit{multi-level cache} reduces communication overhead between HBM and DRAM, and integrating SSDs alleviates the need to load all parameters simultaneously. Lowering both computation and communication demands, along with decreasing dependence on HBM and DRAM for parameter storage, effectively reduces carbon emissions during LLM inference.

\subsection{Dynamic Sparse Mixed-precision LLM Inference}\label{mp_inference}

\begin{wrapfigure}{r}{0.5\textwidth}
  \centering 
  \includegraphics[width=\linewidth]{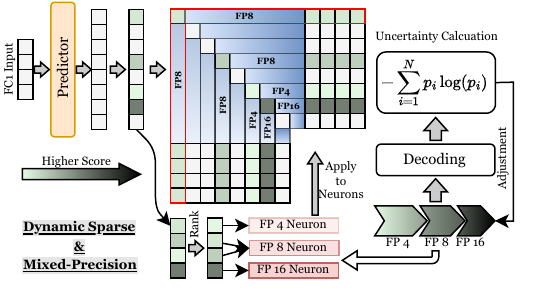}
    \vspace{-6mm}
  \caption{\textit{Dynamic Sparsity and Mixed-Precision.}}
  \label{fig:dynamic_sparse_mq} 
\end{wrapfigure}

\texttt{MP Inference} categorize active neurons into multiple float-point precision types.
The mixed quantization scheme can maintain a certain number of low-precision neurons while keeping critical neurons at high precision.
Therefore, we do not need to prune too many neurons while not all parameters are quantized to a super low bit.
The parameter of FFNs is well compressed, which alleviates the parameter over-correction problem.

As shown in Figure~\ref{fig:dynamic_sparse_mq}, the credential for neuron splitting is the output of the ``Predictor''.
The \texttt{MP Inference} is hinged on one key insight from the Deja Vu predictor~\cite{liu2023deja}, which assigns a predicted score to each neuron.
Neurons with top-k scores, deemed most crucial for the inference, are identified as active.
Based on this observation, neurons with higher scores are loaded in higher float-point precision, while those with lower scores are loaded in lower precision from DRAM.
However, how to explore the adequate neuron ratio under a given fixed memory budget becomes the next problem.

\paragraph{Offline Neuron Ratio Search.} To decide the ratio of different float-point precision neurons for different models.
We design an \textit{uncertainty-guided search method} in Algorithm~\ref{alg:uncertainty_search}.
Given a fixed memory budget, it calculates the decoding precision uncertainty of different ratios of float-point precision neurons.
After that, we decide the ratio of different float-point precision neurons for the current model.
The \texttt{UQEst} we used is defined as:
\begin{equation}
    \texttt{UQEst}(LLM, r_{low}, r_{high})=-\sum_{i>j}^{N}\sum_{k}LLM^{i}_k\log{LLM^{i}_k}.
\end{equation}
Here, $j$ represents the length of the input prompt, $N$ denotes the total length of the sentence, which includes both the prompt and the generated text, and $LLM_{k}^{i}$ is the probability assigned to the $k$-th token in the vocabulary for the $i$-th token generation.
We use the wikitext~\cite{merity2016pointer} for the uncertainty estimation.

\begin{wrapfigure}{r}{0.5\textwidth}
  \centering 
  \vspace{-4mm}
  \includegraphics[width=\linewidth]{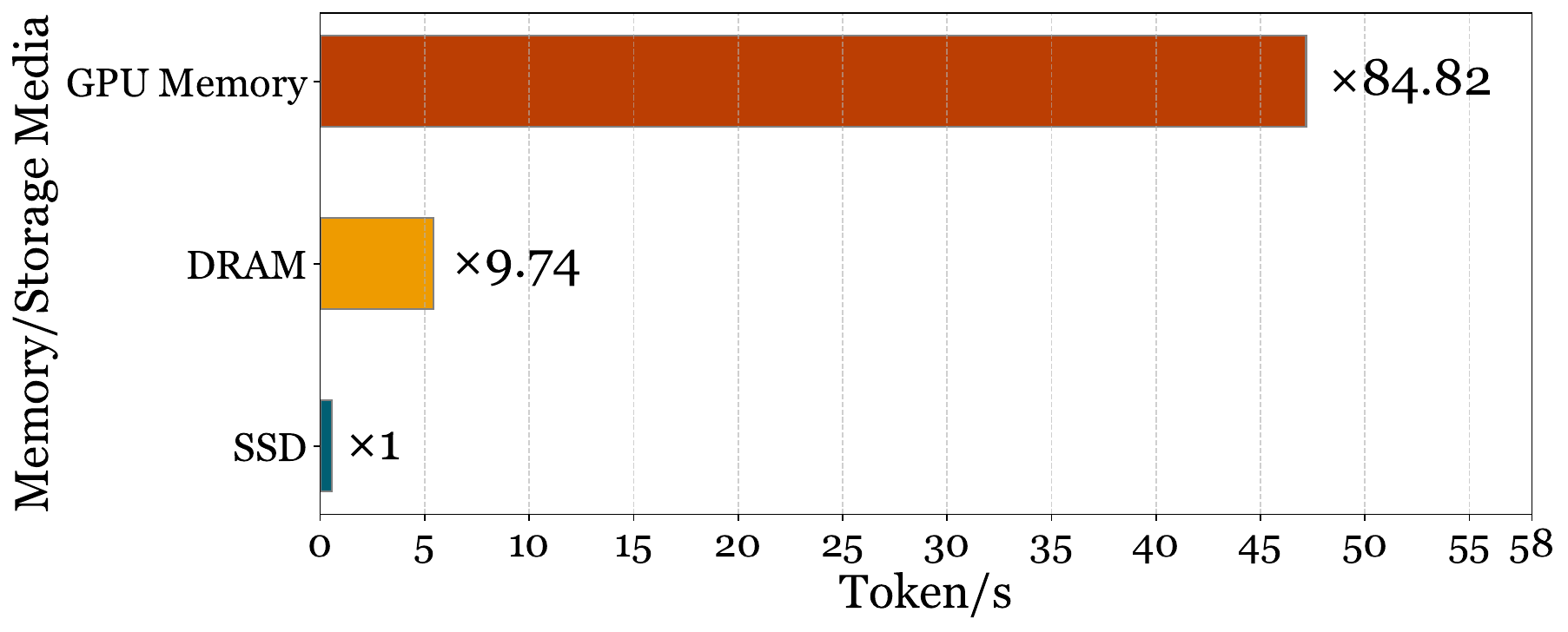}
    \vspace{-6mm}
  \caption{End-to-End Inference Latency on different media (HBM, DRAM and SSD).}
  \vspace{-4mm}
  \label{fig: ete-gpu-dram-ssd} 
\end{wrapfigure}

\begin{algorithm}
\caption{Uncertainty-Guided Search Method}\label{alg:uncertainty_search}
\begin{algorithmic}[1]
\REQUIRE LLM, the target LLM.
\REQUIRE $r_{high}\leftarrow0$, the neuron ratio of high precision.
\REQUIRE $r_{low}$, the neuron ratio of low precision.
\REQUIRE $s$, the search step.
\REQUIRE $n\leftarrow bit(high)/bit(low)$, the number of bit ratio between the high precision and low precision.
\REQUIRE $UQ_{best}\leftarrow+\infty$, the best uncertainty score.
\REQUIRE $R_{ratio}\leftarrow(r_{low}, r_{high})$, the best ratio of neurons.
\WHILE{$r_{low}\geq 0$}
\STATE $r_{high} \leftarrow r_{high} + s$
\STATE $r_{low} \leftarrow r_{low} - s * n$
\STATE $UQ_{score}\leftarrow\texttt{UQEst}(LLM, r_{low}, r_{high})$
\IF{$UQ_{best}\geq UQ_{score}$}
\STATE $UQ_{best}\leftarrow UQ_{score}$
\STATE $R_{ratio}\leftarrow (r_{low}, r_{high})$
\ENDIF
\ENDWHILE
\RETURN $R_{ratio}$
\end{algorithmic}
\end{algorithm}

Compared to using the full parameters of FFNs for inference, \texttt{MP Inference} utilizes only a subset of neurons, which maintains LLM accuracy while using fewer computing resources (lower FLOPS). This reduction in computational demand decreases the carbon emissions associated with LLM inference.
Storing inactive neurons in GPU memory wastes the limited HBM space. By offloading these neurons to DRAM, we conserve GPU memory, further reducing carbon emissions associated with storing LLM parameters in HBM.
More importantly, offloading parameters to another memory device expands the possibilities for advanced host-side cache management. This can further enhance the sustainability and accessibility of LLM inference.

\subsection{High-performance HBM cache management}\label{hbm_cache}

As we employ DRAM to store the model weights, the limited bandwidth and high latency between HBM and DRAM significantly impact inference latency. Figure \ref{fig: ete-gpu-dram-ssd} shows a comparison of inference latency. As shown in Figure \ref{fig: ete-gpu-dram-ssd}, we observe that the inference latency of loading model weights from DRAM is approximately ten times slower than directly caching the model weights in HBM.

There are a number of related studies that design and optimize the cache management for GPU memory \cite{zhang2023_2, liu2024intactkv, nugteren2014detailed, wang2015dacache}. On GPU memory, they usually don't employ dynamic cache designs like LRU, which would cause high overhead of dynamic data swapping between GPU memory and CPU memory (GPU kernels are launched by CPUs) \cite{lin2020pagraph}. One example is demonstrated in LLM-in-a-Flash \cite{alizadeh2023llm}, which uses the Sliding Window Technique to manage the cache loading and eviction. Instead of loading all neuron weights into DRAM, the Sliding Window Technique keeps only the weight rows that are predicted to be necessary for the most recent subset of input tokens. This means that as new tokens are processed, the model only retains the weights relevant to the current and recent tokens, freeing up memory that was previously allocated to older tokens.

\begin{wrapfigure}{r}{0.5\textwidth}
\centering 
\vspace{-4mm}
\includegraphics[width=\linewidth]{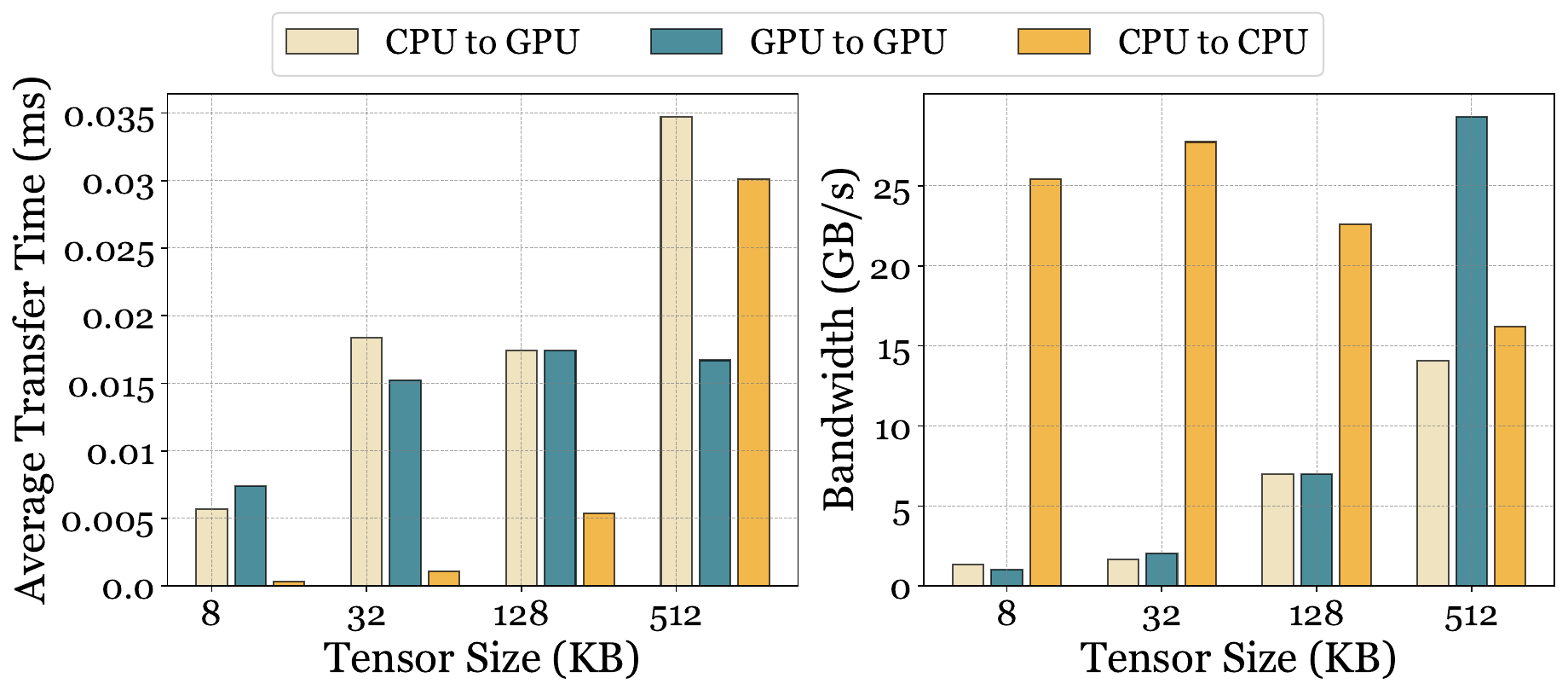}
\vspace{-4mm}
\caption{Transfer time (\textit{left}) and bandwidth (\textit{right}) of different tensor size.}
\label{fig: bandwidth_time_tensor} 
 \end{wrapfigure}

Although the cache design in LLM-in-a-Flash is efficient in DRAM, it is inefficient when we apply them on HBM internally, due to the different copy times and bandwidth between DRAM and HBM at the neuron level. For neuron-level memory copy in GPU memory, it's really inefficient and much slower than in DRAM. Figure \ref{fig: bandwidth_time_tensor} shows the latency and corresponding bandwidth of different sizes of memory copy. From Figure \ref{fig: bandwidth_time_tensor}, we observe that under neuron-level memory copy, HBM is much slower than DRAM, being about 10 times slower. We can also observe that when the data size becomes larger, the copy efficiency of HBM becomes higher than that of DRAM.

\begin{wrapfigure}{r}{0.5\textwidth}
  \centering 
  \includegraphics[width=\linewidth]{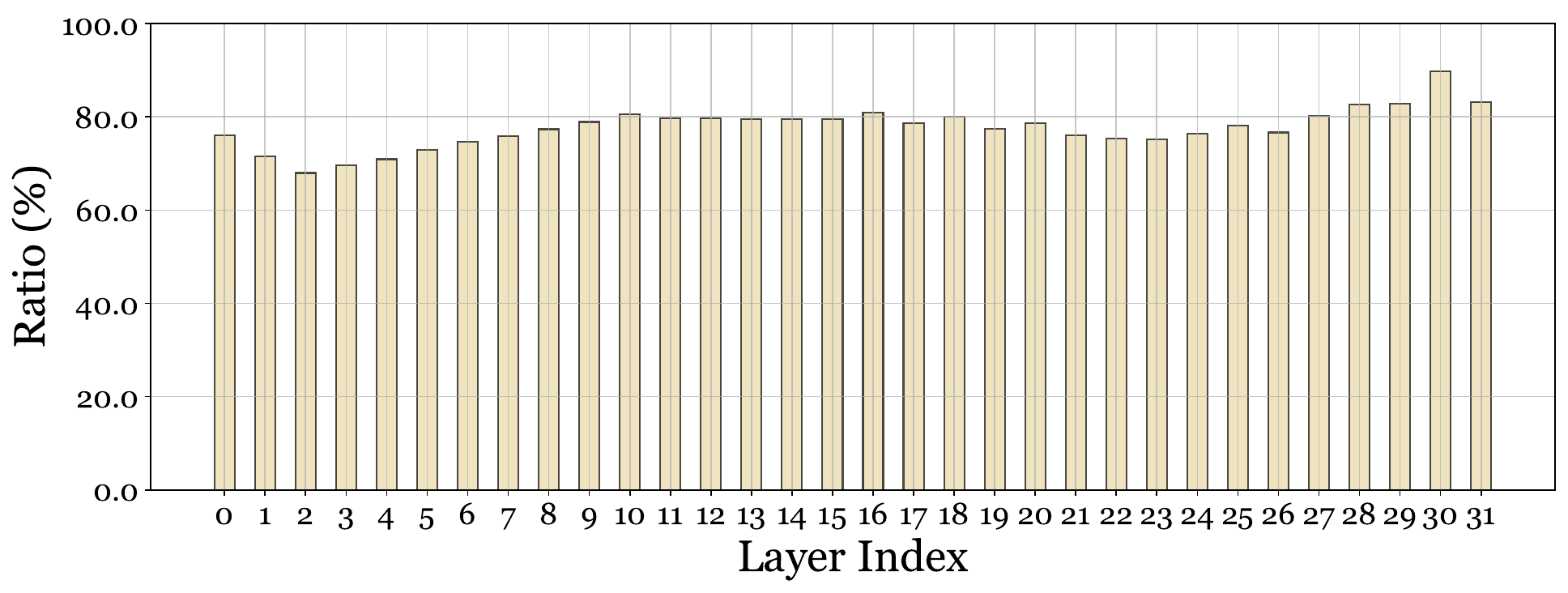}
    \vspace{-4mm}
  \caption{Overlapped neuron ratio between Tokens in different layers. (first half part)}
  \vspace{-4mm}
  \label{fig: overlapped_neurons} 
\end{wrapfigure}

Another key observation is: that there exist overlapping neurons between tokens. Figure \ref{fig: overlapped_neurons} shows the overlapping neurons in each layer and their average ratio. Almost 80\% of the neurons overlap between tokens. Thus, if we can keep these overlapped neurons in the GPU memory and only load the new neurons from DRAM to GPU memory, we can significantly reduce the among of data to be transferred and shorten the latency caused by offloading neurons to DRAM.

\begin{wrapfigure}{r}{0.5\textwidth}
  \centering 
  \includegraphics[width=0.8\linewidth]{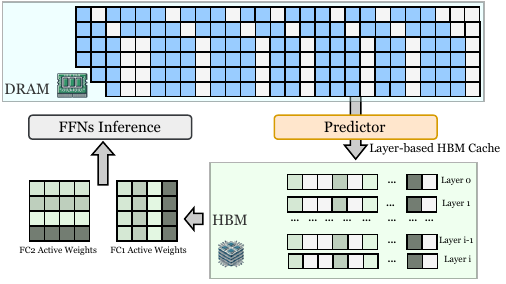}
  \caption{High Performance Layer-based HBM Cache}
  \label{fig: hbm_cache} 
\end{wrapfigure}

\textbf{Overall architecture:} Based on the above observations and analysis, we propose \textit{high-performance layer-based HBM cache}, shown in Figure \ref{fig: hbm_cache}. Specifically, this layer-based cache assigns each layer a \textit{isolated cache unit}. For example, for the LLaMA-2-7B model, which has 32 layers, this HBM cache consists of 32 isolated cache units. In each isolated cache unit, the space is continuous in HBM, and its capacity is allocated based on the number of activated neurons. If the number of activated neurons is $n$ and the size of each neuron is $m$, the capacity of a separate cache is $n * m$. The continuous memory is designed to reduce memory copying overhead when updating the cache, and this continuous memory can be directly used for inference computation, avoiding unnecessary copying from the cache to inference tensors.

\textbf{Cache Policy:} The cache policy is used to update the neurons in each separate cache during inference for different tokens. Here, we employ the \textbf{A}djacent \textbf{T}oken \textbf{U}pdate (\textbf{ATU}) cache policy. ATU only updates the neurons that differ between tokens, and we don't use algorithms like sliding windows proposed by LLM-in-a-Flash or the most widely used LRU \cite{fricker2012versatile} that always retain the hot neurons in the cache. ATU is a trade-off between cache hit ratio and cache management overhead (primarily memory copying overhead). As shown in Figure \ref{fig: overlapped_neurons}, with the proposed high-performance layer-based HBM cache with ATU, the cache hit ratio can reach about 80\%, and the cache management overhead is nearly zero.

\subsection{Pattern-Aware Preloading for SSD}\label{preload_policy}

\begin{wrapfigure}{r}{0.5\textwidth}
\centering 
\vspace{-10mm}
\includegraphics[width=0.9\linewidth]{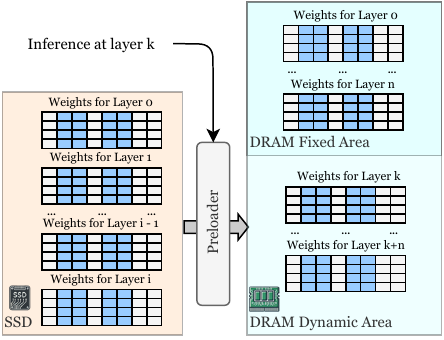}
\caption{\small Pre-loading and Tow-level DRAM Cache}
\vspace{-2mm}
\label{fig: dram_cache} 
\end{wrapfigure}

To address the space limitation of DRAM and improve the overall sustainability, we propose to cache all the model weights in SSD. We designed a flexible and pluggable cache interface for the SSD layer in the proposed multi-level cache, which can be replaced by other flash cache designs including CacheLib \cite{berg2020cachelib}, Kangaroo \cite{mcallister2021kangaroo}, or carbon-sustainable solutions like FairyWREN \cite{298744}. However, the limited bandwidth and high latency between SSD and DRAM will significantly influence the inference latency. As shown in Figure \ref{fig: ete-gpu-dram-ssd}, we observe that the inference latency on SSD (store model weights on SSD and loading the requested neurons to HBM during inference) is approximately 8 times slower than on DRAM and 85 times slower than on HBM.

To efficiently reduce the impact of offloading to SSD, one method is to employ DRAM as the cache tier for SSD and pre-load the model weights to DRAM. There are two solutions for pre-loading: 1) layer-wise; this method directly pre-loads all the neurons of the next few layers from SSD to DRAM in advance. And 2) neuron-level; compared with the layer-wise method, this approach only pre-loads the predicted activated neurons of the next few layers. Generally, these two methods have their pros and cons.

For layer-wise pre-loading, it's simple and effective because it pre-loads all the neurons from SSD to DRAM. There is no complex neuron management, selecting, and memory allocation, and it can loading all the neurons to DRAM without sacrificing the inference accuracy. However, since it pre-loads all the neurons, there exist neurons that are actually not activated, which can waste the loading bandwidth and DRAM space utilizations. For neuron-level pre-loading, it can achieve high memory and bandwidth efficiency since only the predicted activated neurons are identified and loaded from SSD to DRAM. However, it has two key problems. On one hand, it involves complex management. When you load these neurons from SSD to DRAM, you need to map the index of the neurons in the original layer to the address in DRAM. Unlike GPU cache, the DRAM cache capacity is much larger, leading to high memory management overhead. On the other hand, this approach can explicitly influence the prediction accuracy. Although we can use the predictor to estimate the activated neurons of the next several layers based on the current layer, there exist estimation errors that can influence the accuracy \cite{liu2023deja, song2023powerinfer}. For example, when predicting the next one layer, its accuracy is almost 100\%; however, for the next two layers, the accuracy drops to 80\%, and so on. This means you still need to fetch the falsely predicted neurons from SSD during inference, which causes high latency.

Based on the tradeoff analysis of the two schemes mentioned above, we propose \textit{pattern-aware SSD preloading}, as shown in Figure \ref{fig: dram_cache}. It consists of two main modules: 1) preloader, which is used to preload the next a few layers of neurons to be used, load them from the SSD, and insert them into DRAM. And 2) the two-level DRAM cache, which stores and manages the preloaded layers. 

To design a preloader, there are two main factors we need to determine: 1) when to preload the neurons of one layer based on the inference progress such that the loading latency can be hidden, and 2) which neurons in a certain layer should be loaded such that there will be no explicit accuracy impact. First, based on our experiments, the one-layer neuron preloading time (from SSD to DRAM cache) is approximately twice as long as the one layer inference time. Therefore, we only need to preload the neuron from the layer that is two or more layers ahead of the current layer inference. Second, we propose to preload the entire layer to DRAM by identifying the missing neurons in DRAM. 
Second, the two-level DRAM cache consists of two partitions: the fixed area and the dynamic area. The fixed area stores the first $n$ layers of the model. The dynamic area stores the subsequent layers relative to the current layer and changes dynamically during inference based on different layers. The fixed area is used to avoid reloading the first $n$ layers each time inference begins for a new token. The dynamic area is used to avoid reloading layers that have already been inferred.

\subsection{Discussion}
\subsubsection{The Advantage of \mname}

\mname~achieves a better trade-off between carbon emission, inference latency, and available hardware. It utilizes old-fashioned GPUs and low-carbon storage media like available DRAM and SSD for inference. It enables LLM inference on old-fashioned GPUs such as the M40 and 3090 by effectively addressing the limited HBM capacity. It implements a model modularization algorithm that ranks neurons based on their importance, optimizing resource allocation. Moreover, it utilizes dynamic sparse mixed-precision quantization, which reduces both computational demand and communication overhead, improving inference efficiency. Additionally, it introduces a three-level cache system (HBM, DRAM, SSD) that optimizes memory usage and performance. By leveraging older GPUs and improving communication efficiency, \mname~helps reduce operational carbon emissions. It is also worth noting that \mname~is orthogonal to KV cache optimization methods~\cite{zhang2024h2o,zhang2023_2, liu2024intactkv}, as our approach specifically focuses on the offloading of model parameters. By integrating \mname~with KV cache offloading, it is possible to further enhance memory efficiency.

\subsubsection{Limitations}

Despite its advantages,~\mname~has several limitations. First it's still high inference latency. Its reliance on SSDs introduces much higher latency, even with optimizations such as pre-loading and high-performance cache designs. Second, \mname~ can only work for small batch size scenarios. This is due to the Deja Vu predictor; its prediction accuracy is poor under large batch size scenarios, thus \mname~ also performs poorly under large batch sizes.

\section{Experiments}

\subsection{Implementation Details}

\begin{wrapfigure}{r}{0.5\textwidth}
  \centering 
  \vspace{-4mm}
  \includegraphics[width=\linewidth]{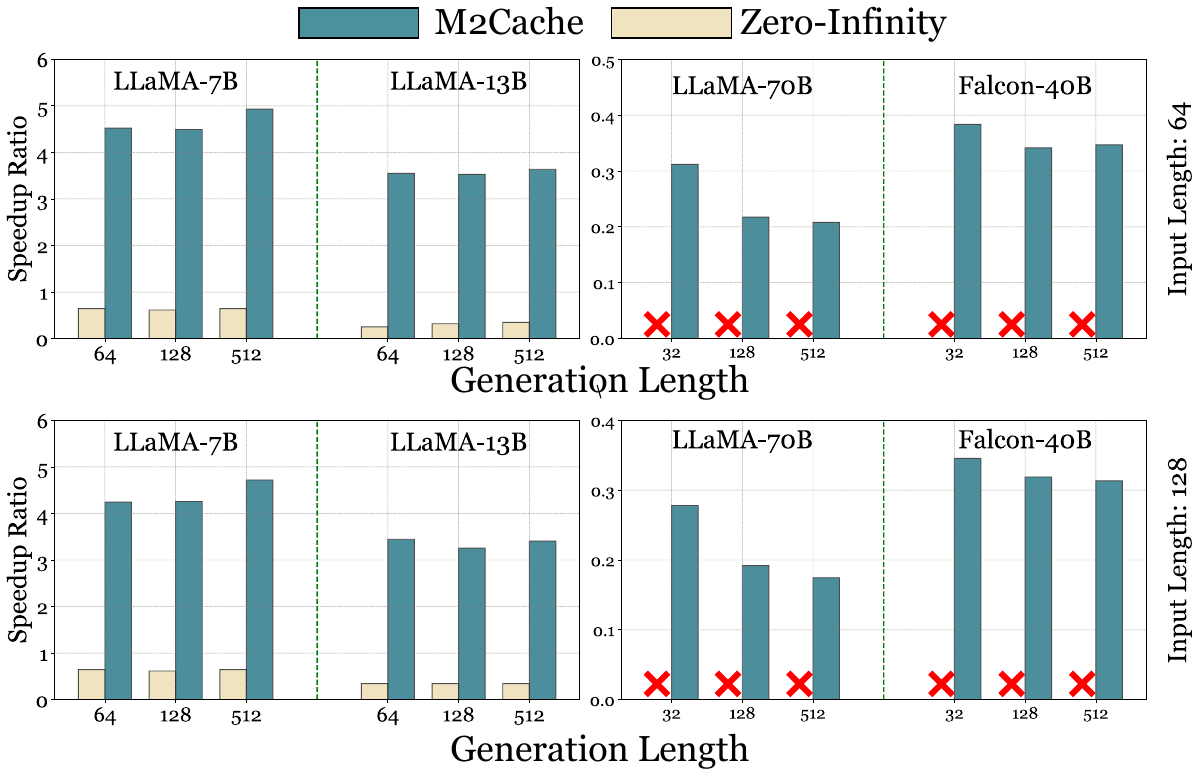}
    \vspace{-4mm}
  \caption{\textit{Generation Speed of various models in FP16 format.} The X-axis indicates the generation text length. The Y axis represents the end-to-end generation speed (token/s). The first row of the figure is configured with an input length of around 64, and the second row with an input length of approximately 128. For LLaMA-70B, the inference speed of ZeRO-Infinity drops to about 0.02 tokens per second. The excessively low generation speed makes using ZeRO-Infinity for inference on models of 70B scale impractical.}
  \vspace{-10mm}
  \label{fig:generatopn_speed} 
\end{wrapfigure}

We implement~\mname~using Pytorch and Python. For the FFN predictors, we adopt the training method from Deja Vu~\cite{liu2023deja}, which includes an adaptive training enhancement.
~\mname~handles model execution by integrating popular large language models (LLMs) like LLaMA-2~\cite{touvron2023llama2} and Falcon~\cite{refinedweb}, utilizing the Transformer architecture~\cite{wolf2020transformers}.
We employ a multi-level cache management system that uses dedicated CUDA streams to transfer data across GPUs and DRAM. 
Additionally, separate I/O threads facilitate the movement of data between the host memory and SSDs, pre-loading layers in advance. 
This setup allows for the overlapping of data transfer with LLM inference calculations. 

\subsection{Experimental Setup}
\textbf{Hardware.} Our inference performance experiments are conducted on a system equipped with a single GeForce RTX 3090 featuring 24GB of HBM, running Ubuntu 22.04. The system configuration also includes 64GB of DRAM, 1TB SSDs, and the CPU is AMD 6950x.

\noindent\textbf{Models.} We evaluate the open-sourced LLaMA-2 model across multiple model sizes—7B, 13B, and 70B—and the Falcon model at 40B. 
Our evaluations employ mixed-precision formats, including FP16, INT8, and INT4, aligning with contemporary practices in LLM research~\cite{liu2023deja, song2023powerinfer}.

\noindent\textbf{Baseline Method.} We benchmark~\mname~against Zero-Infinity~\cite{2022zeroinference}, a state-of-the-art LLM inference framework. 
Although other frameworks like Powerinfer~\cite{song2023powerinfer} are available, they exhibit higher latency under limited CPU computation power. 
Hence, Deepspeed-Infinity, which focuses on GPU-only computing, serves as a more relevant benchmark for our evaluation.

\subsection{End-to-End Evaluations}

We begin our evaluation by comparing the end-to-end inference performance of~\mname~and Zero-Infinity under a typical local deployment setting with a batch size of one~\cite{cai2024medusa}. 
We use prompts sampled from wikitext~\cite{merity2016pointer}, ranging from 64 to 128 sentence length. 
Both~\mname~and Zero-Infinity are tested to generate responses of 64, 128, and 512 tokens for each prompt.

\begin{wrapfigure}{r}{0.5\textwidth}
  \centering 
  \includegraphics[width=\linewidth]{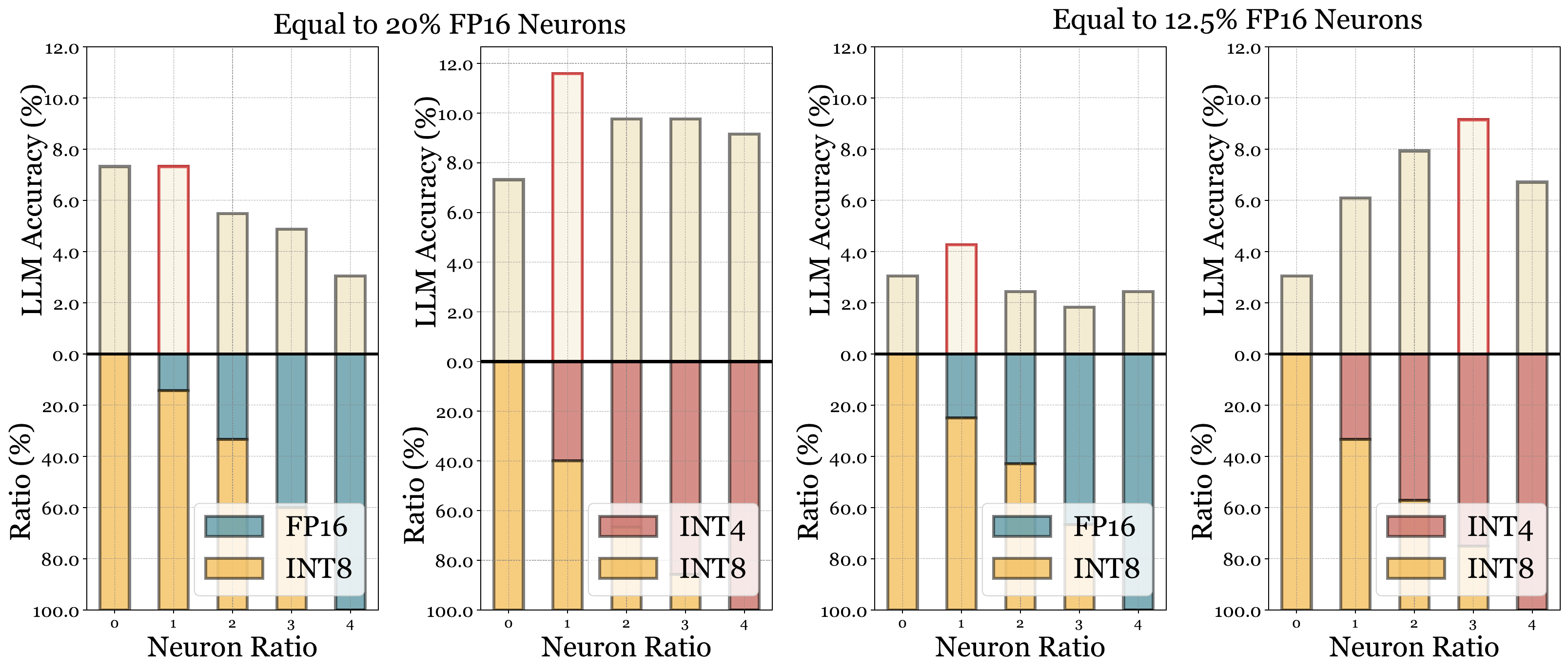}
    \vspace{-4mm}
  \caption{The inference accuracy in HumanEval~\cite{chen2021codex} variety across different ratios of float-point precision. The red box is the ratio of different precision neurons obtained by Algorithm~\ref{alg:uncertainty_search}.
  }
  \label{fig:inference_performance} 
\end{wrapfigure}

Figure~\ref{fig:generatopn_speed} depicts the generation speeds of various models across different input-output configurations. On average,~\mname~outperforms Zero-Infinity with significant speedups, achieving up to 10x faster performance on LLaMA-7B. This advantage of~\mname~is particularly noticeable as the output token count rises, highlighting the increased impact of the generation phase on total inference time.
During this phase,~\mname~utilizes a mix of floating-point precision levels in feed-forward networks (FFNs), which includes 25\% FP16 neurons, 25\% INT8 neurons, and 50\% INT4 neurons, as seen in the LLaMA-13B model. 
This diverse precision strategy reduces communication overheads compared to Zero-Infinity.
Furthermore,~\mname's \texttt{MP Inference} allows for the use of all neurons while consuming only 50\% of the memory required for full-precision neurons. This efficient use of resources underscores the performance benefits of~\mname.

\begin{wrapfigure}{r}{0.5\textwidth}
  \centering 
  \includegraphics[width=0.9\linewidth]{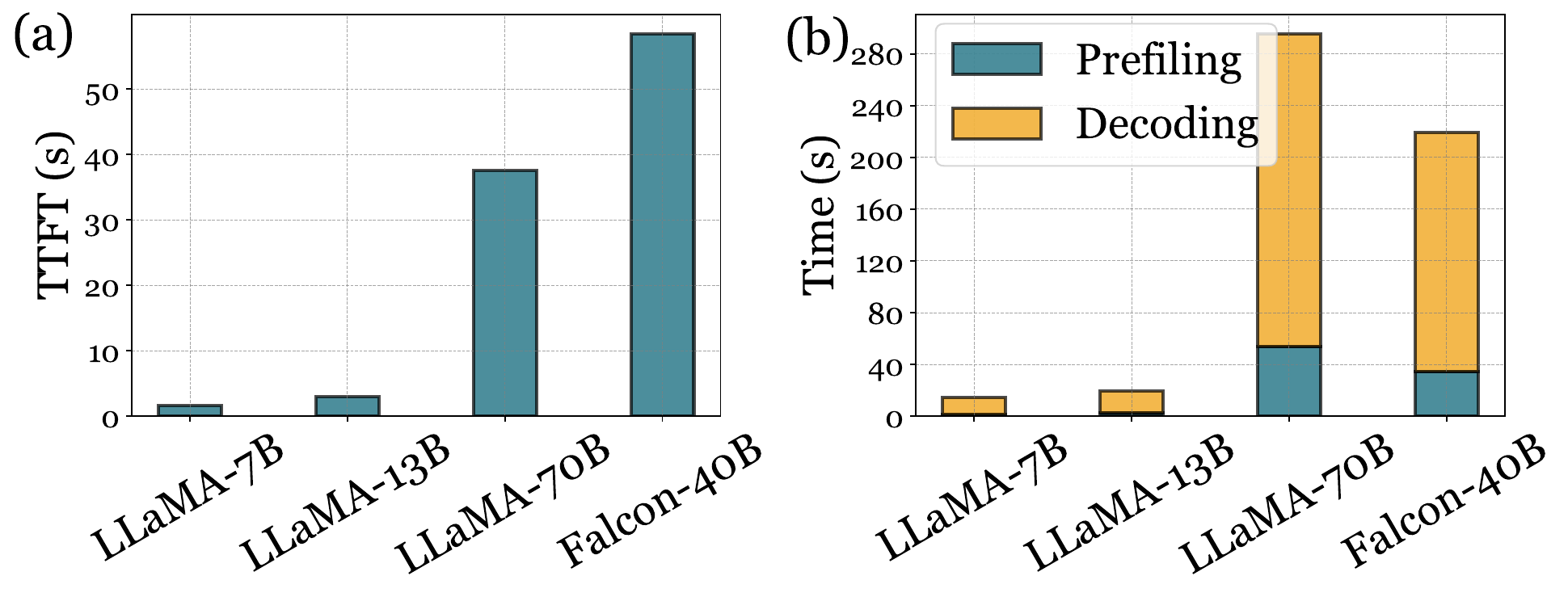}
    \vspace{-4mm}
  \caption{\textit{(a) the time to the first token, (b) the breakdown GPU time.}}
  \label{fig:ttt_breakdown} 
\end{wrapfigure}

The HumanEval requires LLM decoding multiple steps to generate excusable Python code.
Therefore, this task can thoroughly evaluate the effect of different ratios of float-point precision.
Figure~\ref{fig:inference_performance} shows the inference accuracy for various ratios of floating-point neurons. Within the same GPU memory constraints, ~\mname~optimizes the mixed of FP neurons to maximize performance, achieving an average improvement of $2.8\%$ over traditional single-precision inference methods. This enhancement stems from the ability of mixed-precision inference to accommodate more neurons under the same memory budget. The inclusion of lower-precision neurons is crucial, as it can load more neurons under the same memory budgets.
This strategy enables the LLM to maintain high accuracy levels. Furthermore, the neuron ratio determined by Algorithm~\ref{alg:uncertainty_search} outperforms other ratios, demonstrating the effectiveness of the algorithm.

\begin{wrapfigure}{r}{0.5\textwidth}
  \centering 
  \vspace{-4mm}
  \includegraphics[width=0.8\linewidth]{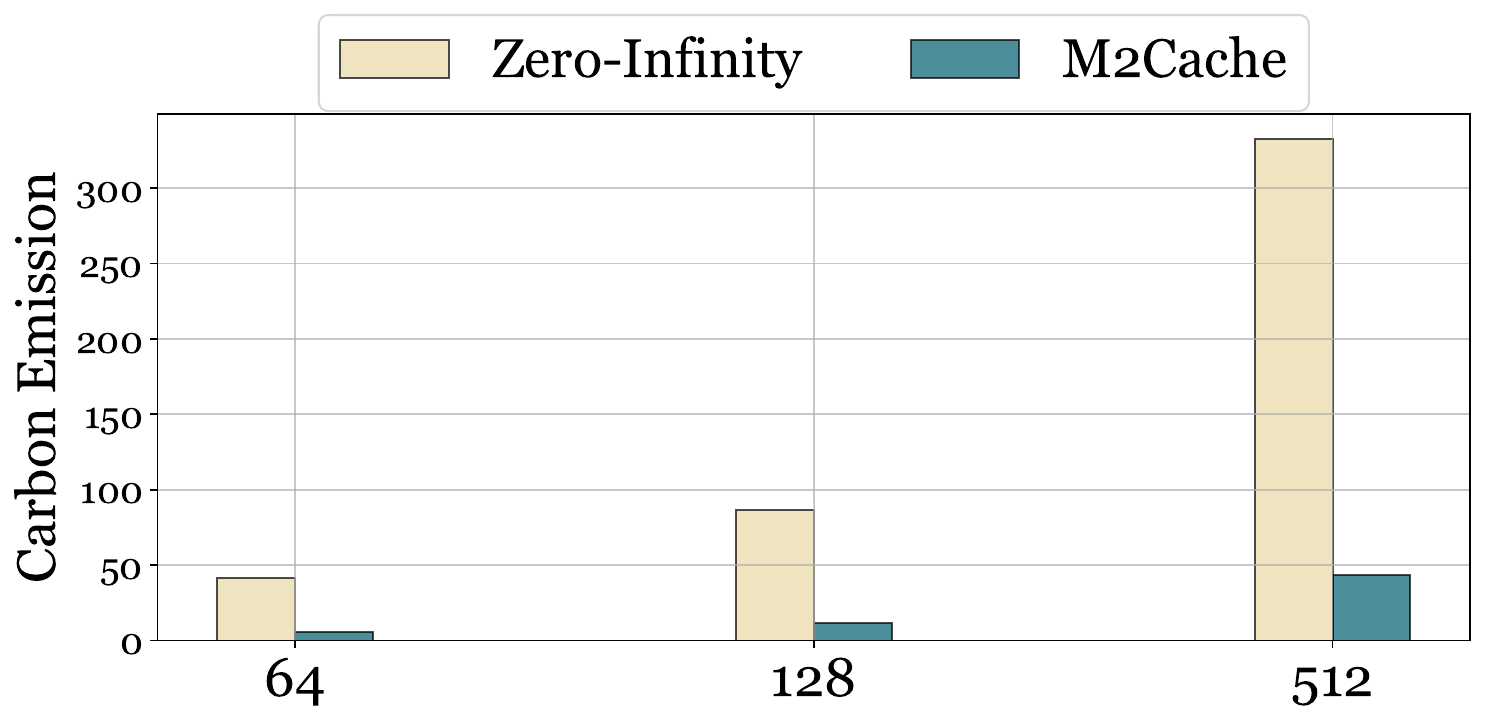}
  \vspace{-2mm}
  \caption{The carbon footprint of~\mname.}
  \vspace{-4mm}
  \label{fig:carbon_emission} 
\end{wrapfigure}

Figure~\ref{fig:ttt_breakdown} displays the time to the first token and the breakdown of GPU time. As the model size increases, the time to generate the first token also rises, while the proportion of decoding time relative to the entire runtime decreases. The time to the first token is particularly higher for the Falcon-40B model due to its unique architectural design.

Figure~\ref{fig:carbon_emission} illustrates the carbon emission reductions achieved by~\mname, ranging from 42 $gCO2$ to 280 $gCO2$. 
These improvements stem from scaling up the model, which leads to the use of more sparse activations and lower precision neurons. 
These changes not only decrease the FLOPs required during decoding but also reduce memory demands on both the GPU and DRAM compared to Zero-Infinity.

\subsection{Ablation Evaluations}

\begin{wrapfigure}{r}{0.5\textwidth}
  \centering 
  \includegraphics[width=\linewidth]{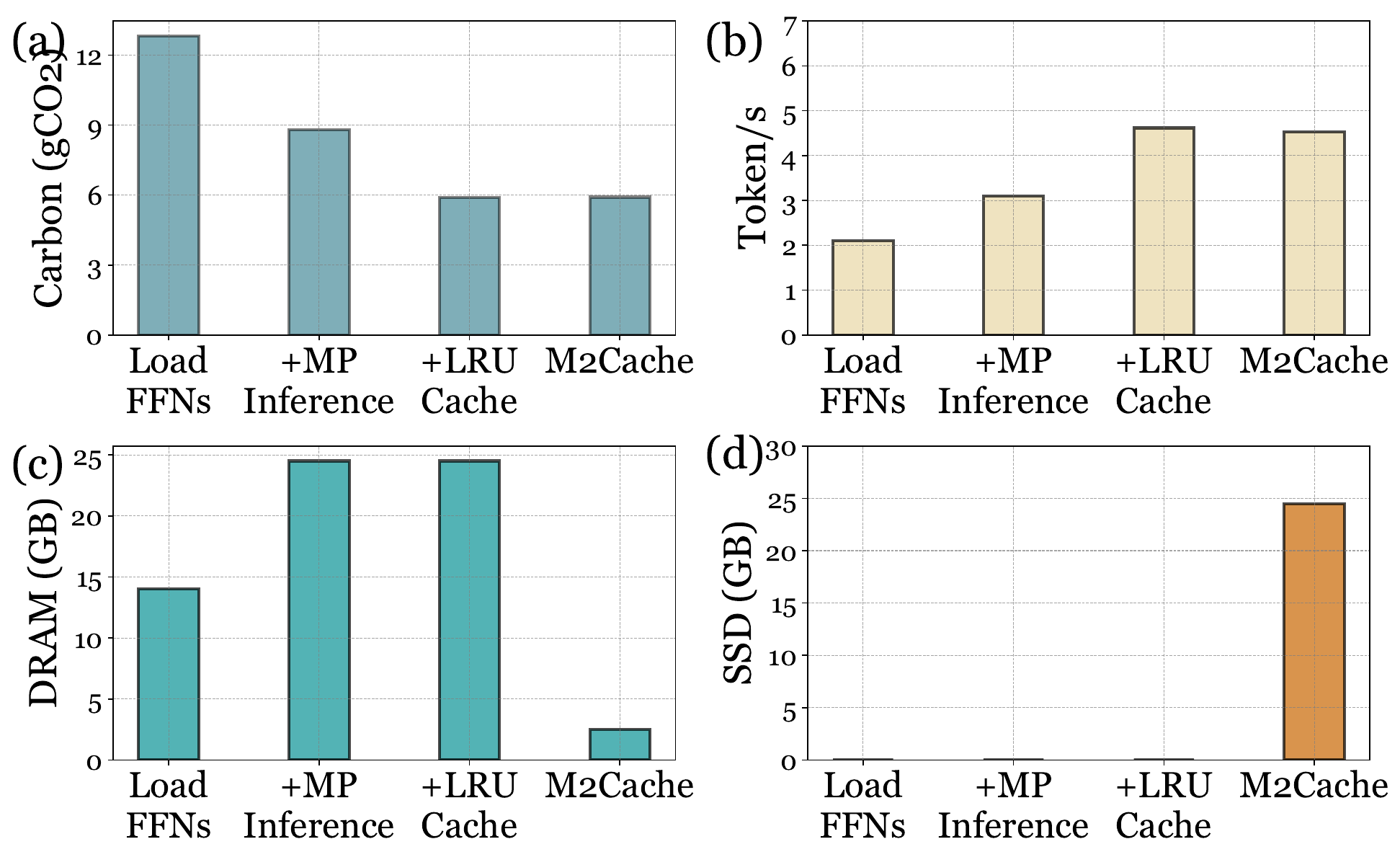}
    \vspace{-4mm}
  \caption{The ablation study for each component of~\mname. The energy consumption of DRAM is 26W for 256GB, and the energy consumption of SSD is 2W \cite{ssds, lee2021greendimm}. The carbon intensity is 820 $gCO2/kWh$ \cite{gupta2022act}.}
  \label{fig:ablation} 
  \vspace{-6mm}
\end{wrapfigure}

\paragraph{Performance Breakdown.} We assess the impact of each component within~\mname~carbon emissions, decoding speed, and GPU and DRAM usage. Figure\ref{fig:ablation} details the contributions of individual~\mname. 
We adopt a step-by-step integration approach to progressively implement~\mname.
Initially, we introduce \texttt{MP Inference} (labeled ``+\texttt{MP Inference}''), which enables dynamic sparse inference with mixed precision neurons. 
However, this stage lacks neuron-level cache management and does not incorporate SSDs.

Building upon ``+\texttt{MP Inference}'', we integrate~\mname's neuron-level cache (denoted ``+LRU Cache''), which utilizes an LRU cache on the GPU to conserve GPU-DRAM bandwidth. 
The final enhancement includes adding SSDs (``+SSDs'') to the ``+LRU Cache'' configuration, demonstrating the full potential of~\mname.

The initial integration of ``+\texttt{MP Inference}'' provides performance boosts around 1 token/s, mainly by reducing the volume of data communicated. 
The addition of ``+LRU Cache'' further increases these gains to 4.62 token/s, leveraging the GPU cache to significantly cut down on data communication volume. 
Finally, the inclusion of ``+SSDs'' achieves reductions of 22 GB in DRAM usage and the carbon emissions remain the same. 
This enhancement is facilitated by our preload policy, which conserves DRAM without compromising inference performance.

\begin{wrapfigure}{r}{0.5\textwidth}
\centering
\vspace{-6mm}
\caption{Comparison of LLM accuracy: HumanEval~\cite{chen2021codex}, which accesses the code-writing capabilities; PIQA~\cite{Bisk2020}, evaluating understanding of commonsense knowledge, RTE~\cite{poliak2020survey}, measuring the major semantic inference; and COPA~\cite{roemmele2011choice} examining commonsense causal reasoning.}
\label{tab:llm_accuracy}
\resizebox{\linewidth}{!}{%
\begin{tabular}{lcccc}
\toprule
 & \textbf{HumanEval} & \textbf{PIQA} & \textbf{RTE} & \textbf{COPA} \\
\midrule
\midrule
LLaMA (ReGLU)-7B & $0.1280$ & $0.7008$ & $0.5343$ & $0.8000$ \\ \midrule
LLaMA (ReGLU)-7B \mname & $0.1159$ & $0.7122$ & $0.5343$ & $0.8000$ \\ \midrule
LLaMA (ReGLU)-13B & $0.1707$ & $0.7106$ & $0.4729$ & $0.6500$ \\ \midrule
LLaMA (ReGLU)-13B \mname & $0.1280$ & $0.6970$ & $0.4729$ & $0.6800$ \\ \midrule
\end{tabular}%
}
\vspace{-4mm}
\end{wrapfigure}

\subsection{LLM Accuracy}

Because~\mname~selectively omits neurons predicted to be inactive and uses mixed floating-point precision, we studied how it affects LLM accuracy. The results, shown in Table~\ref{tab:llm_accuracy}, demonstrate that~\mname~results in only a negligible loss in inference accuracy, regardless of the model size or task type.
While the predictors in each Transformer layer maintain an accuracy rate above 95\%~\cite{liu2023deja}, they occasionally fail to identify some active neurons. 
Consequently, this leads to minor fluctuations in LLM accuracy, which may result in slight decreases or sometimes increases in performance on certain downstream tasks.
Compared to single precision inference, mixed precision generally delivers better performance because it allows for loading more neurons at lower precision

\section{Related Work}

\noindent\textbf{Large Language Models (LLMs)} Since the introduction of Transformer architecture\cite{vaswani2017attention}, it has become the dominant network architecture for language modeling\cite{zhao2023survey, minaee2024large, vaswani2017attention, devlin2018bert, touvron2023llama, touvron2023llama2, dubey2024llama3, jiang2023mistral}. Following the scaling law of transformer-based language models\cite{kaplan2020scaling, hoffmann2022training}, we have seen significant performance improvement by increasing the number of parameters in the model\cite{gpt1, gpt2, gpt3, gpt4, anil2023gemini, reid2024gemini}. Today the number of parameters for most commonly used LLMs ranges from 7B to 65B\cite{touvron2023llama, touvron2023llama2, dubey2024llama3, jiang2023mistral, vicuna2023, bai2023qwen, yang2024qwen2, team2024gemma}, and the weights of these models are stored as 16-bit floating point numbers, which results in a consumption of 14 GB to 130 GB storage just to hold parameters of a model disregarding the storage of KV cache during inference. The amount of consumed storage poses a heavy stress on resource-constrained or old-fashioned computers, making it infeasible to benefit from the performance gain of scaling up number of parameters.

\noindent\textbf{LLM Inference Acceleration.}
LLM inference is an autoregressive process that generates every new token based on previous tokens. The attention mechanism used in LLMs requires $O(n^2)$ computation operations on a sentence of $n$ tokens\cite{vaswani2017attention}, making the inference on GPU-constrained machines slow. To solve this issue, many methods have been proposed\cite{li2024llm, song2023powerinfer, xue2024powerinfer2, zhang2024h2o, kwon2023efficient, mlc-llm}. vLLM\cite{kwon2023efficient}, TensorRT-LLM\cite{nvidia} and TGI\cite{adyenllm} use PagedAttention\cite{kwon2023efficient} to reduce the memory waste in KV cache and achieve flexible sharing of KV cache within and across query requests, leading to higher throughput of LLMs. 
However, these methods don't consider the case when it is unable to hold the entire model in GPU memory, restricting their application on GPU memory-constrained systems.
Powerinfer offloads partial parameters of other storage devices like DRAM and SSDs, reducing the GPU memory requirement for LLM inference.
However, when the CPU resource is limited, the inference efficiency can be poor.

\noindent\textbf{Storage Management for Deep Learning Models.} With the fast progress of deploying deep learning (DL) models in various applications, many research works have proposed methods of storage management for DL models. Quiver\cite{kumar2020quiver} proposed an informed storage cache to improve the throughput of DL training jobs. SHADE\cite{khan2023shade} is a caching system designed for I/O optimizations during distributed deep learning training. To this end, our framework is making pioneering efforts of storage management in LLM inference.

\section{Conclusion}
This paper proposes~\mname, a model cache co-design inference engine can deploy LLM in hardware with limited HBM and DRAM, even the HBM and DRAM together can not load the LLM.
To improve the sustainability and decrease the carbon emission of~\mname, we design the dynamic sparse mixed precision inference and multi-level cache
Extensive experimental results demonstrate that~\mname~significantly decreases the inference speed, and decrease the LLM inference carbon footprint.

\bibliography{ref}




\end{document}